\documentclass[runningheads]{llncs}
\usepackage{graphicx}
\usepackage{amssymb}
\usepackage{multirow}
\usepackage[misc]{ifsym}
\usepackage{xcolor}

\usepackage[T1]{fontenc}
\usepackage{graphicx}
\begin{document}
\title{Dynamic Allocation Hypernetwork with \\Adaptive Model Recalibration for \\Federated Continual Learning}
\titlerunning{FedDAH}
%
\author{Xiaoming Qi\inst{1} \and
Jingyang Zhang \inst{2} \and
Huazhu Fu\inst{3} \and
Guanyu Yang \inst{2} \and
Shuo Li\inst{4} \and
Yueming Jin\inst{1}*}
\authorrunning{XM. Qi et al.}
%
\institute{
Department of Biomedical Engineering and Department of Electrical and Computer Engineering, National University of Singapore, Singapore, Singapore \email{ymjin@nus.edu.sg} 
\and
Key Laboratory of New Generation Artificial Intelligence Technology and Its Interdisciplinary Applications (Southeast University), Ministry of Education, Nanjing, China
 \and
Institute of High Performance Computing, A*STAR  
\and
Departments Biomedical Engineering, and Computer and Data Science, Case Western Reserve University, Cleveland, USA}
\maketitle              
\begin{abstract}
Federated continual learning (FCL) offers an emerging pattern to facilitate the applicability of federated learning (FL) in real-world scenarios, where tasks evolve dynamically and asynchronously across clients, especially in medical scenario. Existing server-side FCL methods in nature domain construct a continually learnable server model by client aggregation on all-involved tasks. However, they are challenged by: (1) Catastrophic forgetting for previously learned tasks, leading to error accumulation in server model, making it difficult to sustain comprehensive knowledge across all tasks. (2) Biased optimization due to asynchronous tasks handled across different clients, leading to the collision of optimization targets of different clients at the same time steps. In this work, we take the first step to propose a novel server-side FCL pattern in medical domain, Dynamic Allocation Hypernetwork with adaptive model recalibration (\textbf{FedDAH}). It is to facilitate collaborative learning under the distinct and dynamic task streams across clients. To alleviate the catastrophic forgetting, we propose a dynamic allocation hypernetwork (DAHyper) where a continually updated hypernetwork is designed to manage the mapping between task identities and their associated model parameters, enabling the dynamic allocation of the model across clients. For the biased optimization, we introduce a novel adaptive model recalibration (AMR) to incorporate the candidate changes of historical models into current server updates, and assign weights to identical tasks across different time steps based on the similarity for continual optimization. Extensive experiments on the AMOS dataset demonstrate the superiority of our FedDAH to other FCL methods on sites with different task streams. The code is available:https://github.com/jinlab-imvr/FedDAH. 

\end{abstract}
\section{Introduction}
\begin{figure*}[!t]
\centering
\includegraphics[width=12 cm]{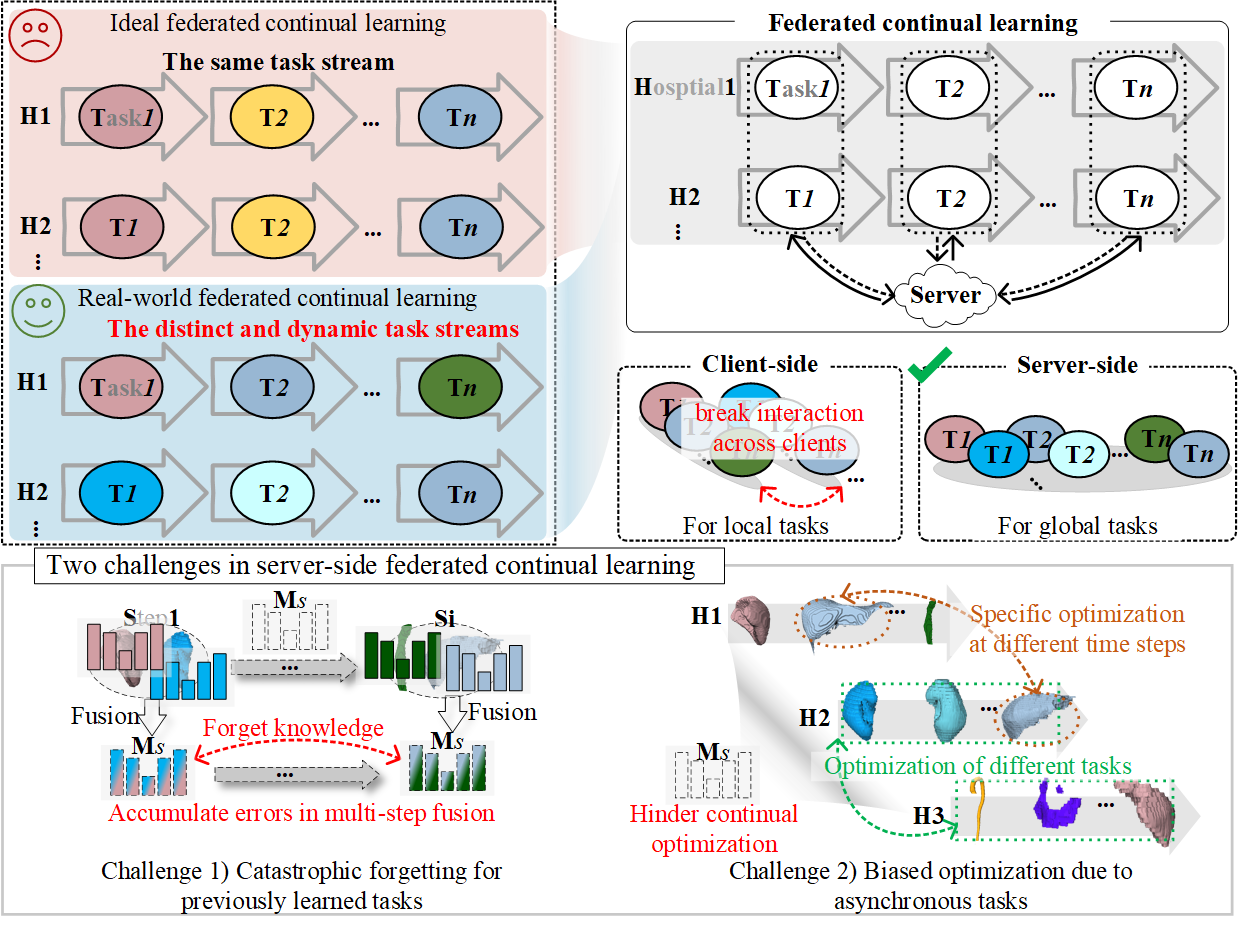}
\caption{\textbf{Task:} Since disease evolves and treatment options change, different clients require to continually evolves on different task orders (asynchronous) or add new tasks (dynamic). \textbf{Challenge:} The construction of a server-side FCL model is challenged by: 1) Catastrophic forgetting for previously learned tasks. 2) Biased optimization due to asynchronous tasks.}
\label{chal}
\end{figure*}

Federated learning (FL) \cite{li2020federated,dong2023federated,qi2022contrastive,zhang2023fedsoda,xu2023federated} is proposed as a paradigm to learn from decentralized data with privacy protection in different clinical centers (clients) and collaboratively learn a global model in server. However, since disease evolves, the development and deployment of treatment options and medical devices occur at varying rates across different clinical centres, this necessitates that clients continuously learn new tasks dynamically and adapt to varying task orders asynchronously \cite{thakur2023self}. These realities limit the applicability of FL in real-world clinical scenarios (Fig.~\ref{chal}). Hence, how to make clients adapt to dynamic and asynchronous task learning, while preserving effective collaborative training, is crucial for facilitating the real-world deployment of the FL model.

To this end, we focus on a more practical FL setting where clients handling dynamic tasks with asynchronous evolution, namely federated continual learning (FCL). Some previous studies propose client-side based methods to meet the challenges in FCL \cite{shenaj2023asynchronous,casado2022concept,yoon2021federated}, which simply employs the off-the-rack continual learning (CL) methods onto client-side updating in federated learning (Fig.~\ref{chal}). However, the client-side FCL ignores the server-side aggregation and breaks the interaction across clients, without effectively utilizing the substantial knowledge available across other clients. Some server-side FCL methods are proposed recently in natural domain \cite{dong2023federated,xu2023federated,criado2022non,ma2022continual}, which aim to construct a continually learnable server model by efficient client aggregation on all-involved tasks. For example, the historical data is utilized to recover the previous optimization by knowledge distillation \cite{zhang2023target,ma2022continual} and consistency constraints \cite{dong2022federated,dong2023federated,ijcai2021p527} in the server fusion process. However, to our best knowledge, the server-side FCL method is still underexplored in medical domain.

Meanwhile, we have identified two main limitations in these existing server-side FCL works:
1) Catastrophic forgetting for previously learned tasks, especially historical data is unavailable for server and future unknown task in FCL. The server accumulates error in FCL and can hardly preserve all task knowledge without data in retraining.
2) Biased optimization due to asynchronous tasks handled across different clients. The existing FCL methods assume each client have the same task order in continual learning. However, the real-world medical sites utilize different task orders in FCL. This leads to the collision of optimization targets of different sites at the same time steps, hindering the provision of an optimal server model for all tasks to each client.

To meet above limitations, one main critical factor lies in how to improve the server memory with harmonious optimization. In this work, our core insight and contribution is to effectively equip the hypernetwork \cite{ha2017hypernetworks} onto the server design to achieve this goal. The idea is motivated by the advantage of the hypernetwork, which can learn an task-specific mapping from a task identity to the task model weights, providing a feasible way to replay all task models of clients to reduce server forgetting and thus facilitate the harmonious optimization. However, there exist some challenges to effectively utilize hypernetwork to tackle FCL problems. For asynchronously evolving tasks in each client, the mapping learning by hypernetwork would be confused with the task-hypernetwork correspondence, misguiding the server optimization. In addition, for server updating, hypernetwork should be recalibrated to update faster for new tasks and slower for existing tasks, which further prompts harmonious optimization.

In this paper, we propose a novel server-side FCL pattern, termed dynamic allocation hypernetwork with adaptive prototype recalibration (\textbf{FedDAH}), aiming to tackle a more realistic collaborative learning setting where distinct and dynamic task streams present in different clients. Specifically, we first propose a \textbf{dynamic allocation hypernetwork (DAHyper)} module. DAHyper presents a continually updated hypernetwork for managing the mapping between task identities and their associated model parameters, enabling dynamic allocation of model parameters across various clients. Through the identity of task I, the hypernetwork is trained to preserve the model parameters of task I. By this setting, the server can establish the mappings between all tasks and model parameters. This enables server updates to leverage all task models learned by clients without accumulating errors. We further design an \textbf{adaptive model recalibration (AMR)}. Benefiting from the defined mapping mechanism between task and model parameter, the server could obtain the same task parameter in the asynchronous tasks. AMR assigns a calibration to each model optimization based on the contrastive similarity, enabling rapid integration of new knowledge from new task models while retaining previously learned knowledge with less fading. We have conducted extensive experiments on AMOS dataset for abdominal organ segmentation with multi-center, multi-vendor, multi-modality, multi-phase, multi-disease patients. Our FedDAH achieves the substantial improvement compared with the state-of-the-art methods. 

Overall, our contributions can be summarized as follows:
\begin{enumerate}
    \item For the first time, we propose a novel server-side FCL pattern in the medical scenario, FedDAH, to tackle a more practical collaborative training setting where different clinical clients have their distinct and dynamic task streams. 
    \item A novel server-side model aggregation pattern, DAHyper, is proposed to manage and allocate the model parameters across various clients without error accumulation caused by forgetting in FCL.
    \item A novel server-side model optimization strategy, AMR, is proposed to calibrate the continual optimization on asynchronous task streams in FCL.
\end{enumerate}

\section{Methodology}
\begin{figure*}[!t]
\centering
\includegraphics[width=12 cm]{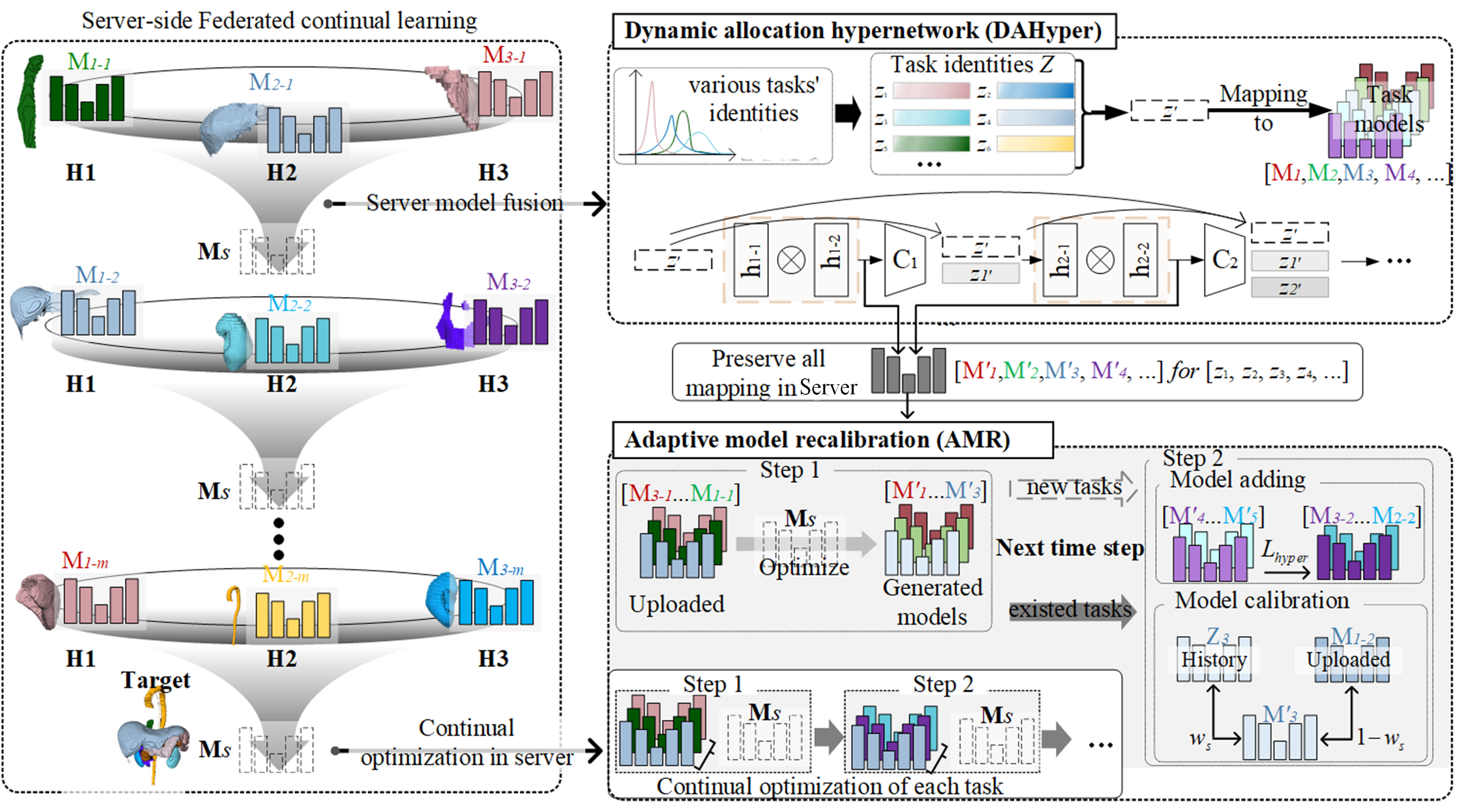}
\caption{The framework of FedDAH: (a) Dynamic allocation hypernetwork preserves the mappings (task identity to model weights) by the hypernetwork to avoid knowledge forgetting. (b) Adaptive model recalibration assigns a calibration based on the contrastive similarity for continual optimization on asynchronous tasks.}
\label{network}
\end{figure*}

We propose a novel server-side FCL framework FedDAH, aiming to tackle the crucial yet challenging scenario that different clients have different task streams (Fig.~\ref{network}). FedDAH consists of (1) DAHyper, which is to preserve the mappings of task identities to model weights, to avoid knowledge forgetting and allocate a required model to the client (Sec. 2.1); (2) AMR, which is to assign a calibration to each model optimization for continual optimization on the distinct task streams (Sec. 2.2).

\subsection{DAHyper for knowledge preservation}
In our FedDAH, DAHyper defines a novel hypernetwork to generate the whole model parameter from task identities for each client knowledge preservation and away from catastrophic forgetting in FCL. It contains (1) Task identity definition and (2) Hypernetwork construction. The details are as follows:

\vspace{1mm}
\noindent \textbf{Rationale:} In CL, a neural network $f(x,\theta)$ with weights $\theta$ is given from a set of tasks $\{( X_{1}, Y_{1}) ,..., $ $(X_{T}, Y_{T})  \} $. Instead of retain $f(x,\theta)$ for previous tasks in continual learning, a metamodel $f_{h}(e, \theta _{h})$ (DAHyper) maps a task embedding $e$ to weights $\theta$ by weights $\theta_{h}$. Through training $f_{h}$ on the acquired input(task embedding $e$)-output(weights $\theta$) mappings, all the task knowledge can be preserved. Hence, hypernetworks can address catastrophic forgetting in continual learning at the meta level. Different from the generation of one layer parameter in traditional hypernetwork\cite{von2020continual}, our DAHyper enables to generate the weights for the entire network, and learns the parameters $\theta _{h}$ of a metamodel to output the model parameter $\theta$ for a specific task. 

\vspace{1mm}
\noindent \textbf{Task identity definition:} Considering the dynamically updated tasks in FCL, DAHyper proposes a unique pattern to distinguish different tasks and associate tasks with the corresponding model weights. In the traditional hypernetwork, a task embedding $e$ is randomly generated during training for a layer's weight. Since $e$ is random value for a layer, this pattern cannot be applied to FCL to distinguish various tasks. In DAHyper, we define the task identity set $Z=\{z_{1}, z_{2}, ...\}$ for various tasks in continual learning process. The element of $Z$ is a vector generated from normal distribution ($N(\mu_{z},\sigma^{2})$) with different $\mu_{z}$ and $\sigma$. Considering the different tasks in different clients, DAHyper designs $z_{i}$ from $Z$ for each task to distinguish the different tasks in server.

\vspace{1mm}
\noindent \textbf{Hypernetwork construction:} Since DAHyper generates the entire model weights of Task $i$, each layer parameter of the model require to be considered. For Task $i$ model each layer, the parameters of a layer are associated with the task identity $z_{i}$ and the previous layers. Hence, DAHyper defines the hypernetwork according to the task identity and inter-layer consistency.

\textit{In task identity:} We assume the parameters of a layer $j$ in the Task $i$ model are stored in a matrix $K^{j} \in  \mathbb{R}^{N_{in}f_{s}\times N_{out}f_{s}}$, where $f_{s}\times f_{s}$, $N_{in}$, and $N_{out}$ are the filter sizes, input size, and output size of the layer.Since the $K^{j}$ can be viewed as $N_{in}$ slices of a matrix $K^{j}_{in}$ with $f_{s}\times N_{out}f_{s}$, we generate the parameter by two-layer linear network. In the first layer ($h_{1-1}$), the $z_{i}$ is projected into the $N_{in}$ vectors $a_i$, with $N_{in}$ different matrices $W_{i} \in \mathbb{R}^{d\times N_{z}}$ and bias $B_{i} \in \mathbb{R}^{d}$, where $d$ and $N_{z}$ are the size of the hidden layer and $z_{i}$. The $h_{1-2}$ takes the vector $a_{i}$ and projects it into $K^{j}_{in}$ using weights $W_{o}\in \mathbb{R}^{f_{s}\times N_{out}f_{s}\times d}$ and $B_{o}\in \mathbb{R}^{f_{s}\times N_{out}f_{s}}$. The $K^{j}$ is a concatenation of every $K^{j}_{in}$. The whole process can be expressed as:
\begin{equation}
a_{i} = W_{i}z_{i}+B_{i}, ~~
K^{j}_{in}=W_{o}a_{i}+B_{o}, ~~
K^{j} = Concat(K^{j}_{1},...,K^{j}_{N_{in}})
\end{equation}

\textit{In inter-layer consistency:} The next layer parameters are not only associated with the task identity $z_{i}$, but also keep the inter-layer consistency with the previous layer parameters $K^{j}$. According to this, DAHyper introduces a mechanism: Firstly, the previous layer parameters $K^{j}$ is encoder into a vector $z1'$ with the same size of $z_{i}$ by Encoder $C_{1}$. Then the concatenation of $z_{i}$ and $z1'$ is the input of next layer generation ($h_{2-1} \& h_{2-2}$). The feedforward process of $h_{2-1} \& h_{2-2}$ is the same as $h_{1-1} \& h_{1-2}$. Following this operation, the concatenation of $z_{i}$ and $z1'$ also will be concatenated with the further more outputs ($\{z2', z3',...\}$). Finally, DAHyper can obtain the parameters $\theta$ of model $M_{i}'$ in server for Task $i$ based on $z_{i}$.

\subsection{AMR for continual optimization}
To avoid the optimization bias caused by asynchronous tasks in FCL, AMR treats the first model weights for each task as a basic model (standard) and ensures continual optimization of each basic model by calculating a calibration based on the similarity to the same model weights uploaded at different time steps in FCL. 

AMR ensures the continual optimization from two aspects: (1) Continual optimization of different tasks. For the uploaded different tasks, AMR defines the historical calibration to regularize each task in server. (2) Continual optimization of the same task. For the models with the same task yet uploaded at different time steps, AMR defines the similarity among the models and utilizes the similarity as weights to guide optimization.

\vspace{1mm}
\noindent \textbf{Continual optimization on different tasks.} For the uploaded models of different tasks in server, AMR requires to optimize the DAHyper with the models and balance the optimization on the current tasks and previous tasks in different time steps. Hence, AMR treats each task model as a basic model, and the optimization of each basic model during the following steps should not be degraded by other basic models. AMR takes a two-stage learning ($\mathcal{L}_{hyper}$) on the current task and historical basic models. Firstly, a candidate change $\bigtriangleup \theta _{h}$ is calculated by minimizing the loss on the current task $\mathcal{L}_{task} (\theta _{h}, z_{i}, M_{i})$, where $\theta _{h}$, $z_{i}$, and $M_{i}$ are the parameters of DAHyper, task identity, and target model of the current task $i$. Through the $\mathcal{L}_{task}$, we can guide the DAHyper to obtain the $\theta$ for each task. Here, AMR utilizes L2 distance to calculate $\mathcal{L}_{task}$. Secondly, AMR regularizes the historical basic models while attempting to learn the current task by:  
\begin{equation}
\mathcal{L}_{R} = \frac{1}{T-1} \sum_{t=1}^{T-1} ||f_{h}(z_{t},\theta_{h}^{*}) - f_{h}(z_{t},\theta_{h}+\bigtriangleup \theta _{h})||^{2},
\end{equation}
where $z_{t}$ and $\theta_{h}^{*}$ are the task identity of task $t$ and the set of DAHyper parameters before attempting to learn task $T$ (current task). Since the knowledge of historical basic models is preserved by DAHyper without current task optimizations, the regularization $\mathcal{L}_{R}$ takes the minimization of difference between updated output and historical knowledge to ensure the DAHyper effective on different basic models (current and previous tasks) at the same time. Hence, the $\mathcal{L}_{hyper} = \mathcal{L}_{task}+\beta \mathcal{L}_{R}$. The $\beta$ is a hyperparameter of $\mathcal{L}_{R}$. 

\vspace{1mm}
\noindent \textbf{Continual optimization of the same task.} Besides the $\mathcal{L}_{hyper}$ controls the optimization on different basic models, the basic models for the same task at different time steps in FCL also require continual optimization. Hence, we further develop a recalibration based on $\mathcal{L}_{hyper}$. The process is shown in Fig.~\ref{network}: (1) In the step 1, the weight parameters $M_{1}',M_{2}',M_{3}'$ generated by DAHyper are optimized by the uploaded models $M_{1-1},M_{2-1},M_{3-1}$ from different clients. (2) Then, the optimized $M_{1}, M_{2}, M_{3}$ are treated as 3 basic models in server. (3) In the step 2, server receives $M_{1-2},M_{2-2},M_{3-2}$ from H1, H2, and H3. $M_{1-2}$ is correspond to the existing basic model $M_{3}$. With the new $M_{3}'$ generated by DAHyper, there are 2 optimization targets ($M_{3}$ and $M_{1-2}$). Considering the convergence of $M_{1-2}$ worse than the historical model $M_{3}$, AMR takes the $M_{1-2}$ as regularization to benefit $M_{3}$. Hence, AMR calculates the similarity weights of $W_{s}(M_{3}',M_{3})$ as the basic, and utilize $(1-W_{s})(M_{3}',M_{1-2})$ as further recalibration. The similarity is measured by JS divergence\cite{fuglede2004jensen}. 

According to the weights, the final loss is: 
\begin{equation}
\mathcal{L} = W_{s} [\mathcal{L}_{task}(M_{3}',M_{3}) +\beta _{1} \mathcal{L}_{R}1] + (1-W_{s}) [ \mathcal{L}_{task}(M_{3}',M_{1-2}) +\beta _{2} \mathcal{L}_{R}2].
\end{equation}
Through treating the new updated model of existing basic model as the recalibration, AMR ensures the continual optimization of the same task at different time steps. 

\section{Experiments and Results}
\subsection{Dataset and Implementation} 
\noindent \textbf{Dataset and Evaluation Metric:} To evaluate the performance of our FedDAH, we conduct experiments on the AMOS dataset \cite{ji2022amos}. AMOS provides 500 CT scans collected from multi-center, multi-vendor, multi-modality, multi-phase, multi-disease patients, each with voxel-level segmentation annotations of 15 abdominal organs. We reconstruct the AMOS dataset to simulate a more realistic clinical FCL. We set 4 clients (C1-C4) with each having 125 CT respectively. The 125 CT are divided into the training and testing sets as 4:1. Each client takes all the 15 organs for testing. Considering the high likelihood that different clinical centers may have some identical tasks at the beginning, we select some organ segmentation as the initialization task existing in all clients (left kidney and right kidney in this work). This can also evaluate the effectiveness of FCL methods on the same task streams.  In addition, different clinical centers are likely to tackle the same or varying tasks in differing sequences in the upcoming steps. We further divided other organs into shared and unique parts to evaluate FCL methods on the same tasks with different streams, and on distinct tasks with different streams. Each client conducts continual learning by a random order based on the combination of the shared part and the unique part. Details are shown in Tab.~\ref{dataset}. 
We employ Dice similarity coefficient \cite{dice1945measures} as the evaluation metric for this segmentation task. We calculate the average Dice of all organs in the continual learning process for a fair comparison.

\begin{table}[!t]
\centering
\caption{The details of the dataset and the settings of each client in FCL.}
\scalebox{0.9}{
\begin{tabular}{ccccc}
\hline
\multirow{2}{*}{Clients} & \multirow{2}{*}{Num} & \multirow{2}{*}{Task 1} & \multicolumn{2}{c}{Task 2-8 (random order)}  \\ \cline{4-5} 
                         &                         &                                   & Shared          & Unique                    \\ \hline
C1                       & 125                     &                                   & spleen,          & bladder, prostate         \\ \cline{5-5} 
C2                       & 125                     & left kidney,                      & stomach,pancreas, & aorta, inferior vena cava \\ \cline{5-5} 
C3                       & 125                     & right kidney                      & gallbladder,     & duodenum, esophagus       \\ \cline{5-5} 
C4                       & 125                     &                                   & liver            & left, right adrenal gland \\ \hline
\end{tabular}}
\label{dataset}
\end{table}

\vspace{1mm}
\noindent \textbf{Implementation:} Our FedDAH takes 3D Unet \cite{cciccek20163d} as the basic segmentation network for each client's continual learning. In each client training, the network is based on Pytorch with the learning rate of $1\times 10^{-3}$, Adam \cite{adam} optimizer, and a batch size of 1. The communication of FL is conducted after every $E=5$ in client training until $T=20$ in total for each task. In each client training, data augmentation (rotation, translation, scale, and mirror) and maximum connected domain are the post-processing. In server, the hypernetwork is optimized by Adam with the learning rate of $1\times 10^{-3}$. All experiments are performed on four NVIDIA A6000 GPUs. 

\begin{table}[!t]
\centering
\caption{The mean Dice score of each client evaluates the superior ability of continual learning in FedDAH (the testing of each method is performed on 15 organs).}
\scalebox{0.9}{
\begin{tabular}{ccccccccccc}
\hline
Method & FedAvg                   & FBL                      & FedWeIT                  & FedSpace                 & \multicolumn{4}{c}{FedDAH}                       & \multirow{2}{*}{Local} & \multirow{2}{*}{Centralized} \\ \cline{1-1} \cline{6-9}
client & \cite{mcmahan2017communication} & \cite{dong2023federated} & \cite{yoon2021federated} & \cite{shenaj2023asynchronous} & -DAHyper & -$\mathcal{L}_{R}$ & -$W_{s}$ & Full   &   &                       \\ \hline
C1     & 0.019                    & 0.213                   & 0.700                    & 0.763                   & 0.682                   & 0.347         & 0.432     & 0.801  & 0.667                        & 0.831                \\
C2     & 0.020                    & 0.255                   & 0.723                    & 0.761                   & 0.711                   &0.338          &0.466      & 0.805 &  0.631                      & 0.801                 \\
C3     & 0.018                    & 0.236                   & 0.707                    & 0.733                   & 0.679                   &0.340          & 0.458     & 0.812  & 0.589                       & 0.828                  \\
C4     & 0.016                    & 0.131                   & 0.659                    & 0.744                   & 0.708                    &0.283          &0.414      & 0.807 &  0.577                      & 0.820                 \\ \hline
\end{tabular}}
\label{Compare}
\end{table}

\subsection{Experimental Results} 
\noindent \textbf{Quantitative and Qualitative Analysis.}
We evaluate our FedDAH from quantitative and qualitative aspects by the comparison with the state-of-the-art FCL methods (FBL \cite{dong2023federated}, FedWeIT \cite{yoon2021federated}, and FedSpace \cite{shenaj2023asynchronous}) and CL based on FedAvg \cite{mcmahan2017communication}. FBL utilizes task relations to benefit different clients' optimization to realize the continual learning at each client. FedWeIT designs knowledge distillation at client optimization to benefit different tasks in continual learning. FedSpace benefits different tasks in continual learning by additional task data. The CL based on FedAvg directly utilize continual learning setting at each client and take FedAvg to realize FL setting.
Apart from these, we also centralize the training data for model optimization as the upper bound, and just train the local models by the local data with all organ labels. The results are shown in Tab.~\ref{Compare}. 

We can see that (1) Our FedDAH achieves the best mean Dice compared with others, peaking at 0.801, 0.805, 0.812, and 0.807 Dice for the four clients. This indicates that our FedDAH can alleviate knowledge forgetting and asynchronous server optimization difficult under this more realistic FCL setting, which different clinical centres have different task streams. (2) FedAvg and FBL show the worse performance than others, showing that directly combining CL with conventional FL method still struggle to tackle the challenges brought by different task streams in FCL.
(3) Compared with the centralized training and local training methods, FedDAH achieves better performance than local training model and comparable performance with centralized training model. This indicates that the FL can improve the different client model optimization with different tasks in the real-world by FedDAH. (4) Through the comparison of different clients, the local training could achieve a better performance than FedAvg and FBL. This is caused by the challenges of catastrophic forgetting and asynchronous tasks in FCL. However, the model sharing technology in traditional CL methods hardly overcome these challenges.
In FedWeIT and FedSpace, the client localization for each task in client optimization requires additional optimizations in clients and worse than centralized training. This results also indicates that our FedDAH could alleviate the challenges of catastrophic forgetting and asynchronous task streams in FCL.

\begin{figure*}[!t]
\centering
\includegraphics[width=12cm]{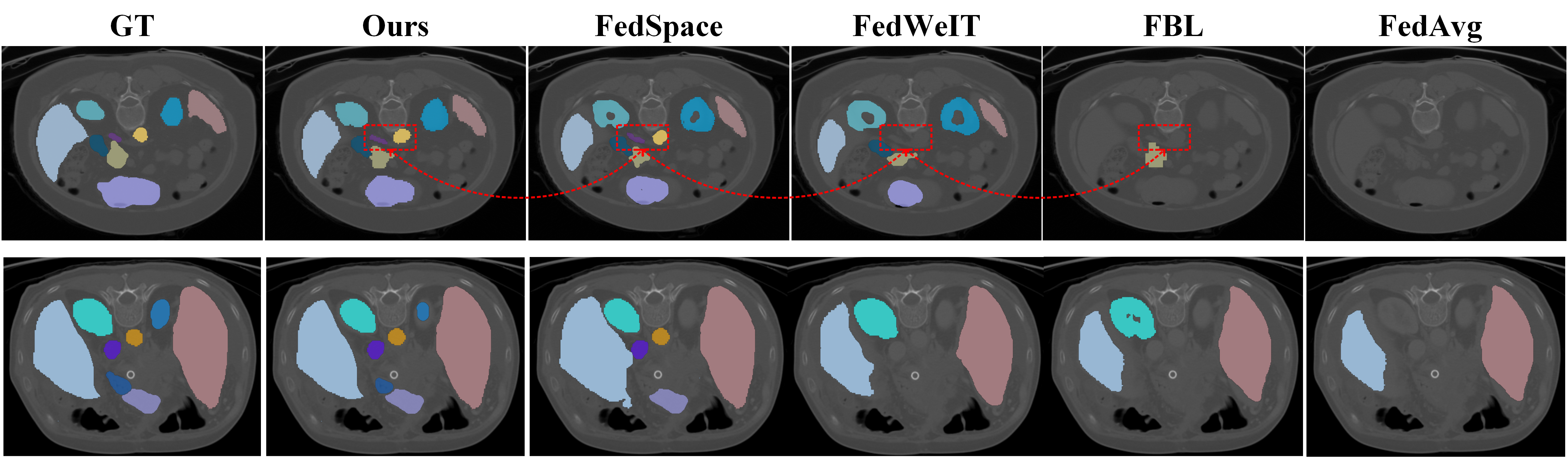}
\caption{The visual results indicate the superior performance of FedDAH. Especially in the red box, we show a task optimized by other clients, and our FedDAH provides more complete segmentation.}
\label{vis}
\end{figure*}

To further evaluate the performance, we visualized the segmentation of different methods. From the visual results in Fig.~\ref{vis}, it can be found that our FedDAH could achieve the accurate and complete segmentation masks during the continual learning process. Through the comparison of the same regions across different methods, we can see that the existing FL and FCL methods tend to omit some regions which are difficult to segment due to catastrophic forgetting.

\vspace{1mm}
\noindent \textbf{Ablation Study.} To evaluate the contributions of each part in FedDAH, we design different ablation studies based on the following experimental settings. -DAHyper: we remove the DAHyper module to evaluate the effectiveness on overcoming catastrophic forgetting in FCL. -$\mathcal{L}_{R}$: we remove the historical calibration to evaluate the effectiveness of history in continual optimization of different tasks in FCL. -$W_{s}$: we remove the similarity weight to evaluate the effectiveness of similarity calibration on the continual optimization of the same task. 

\begin{figure*}[!t]
\centering
\includegraphics[width=12cm]{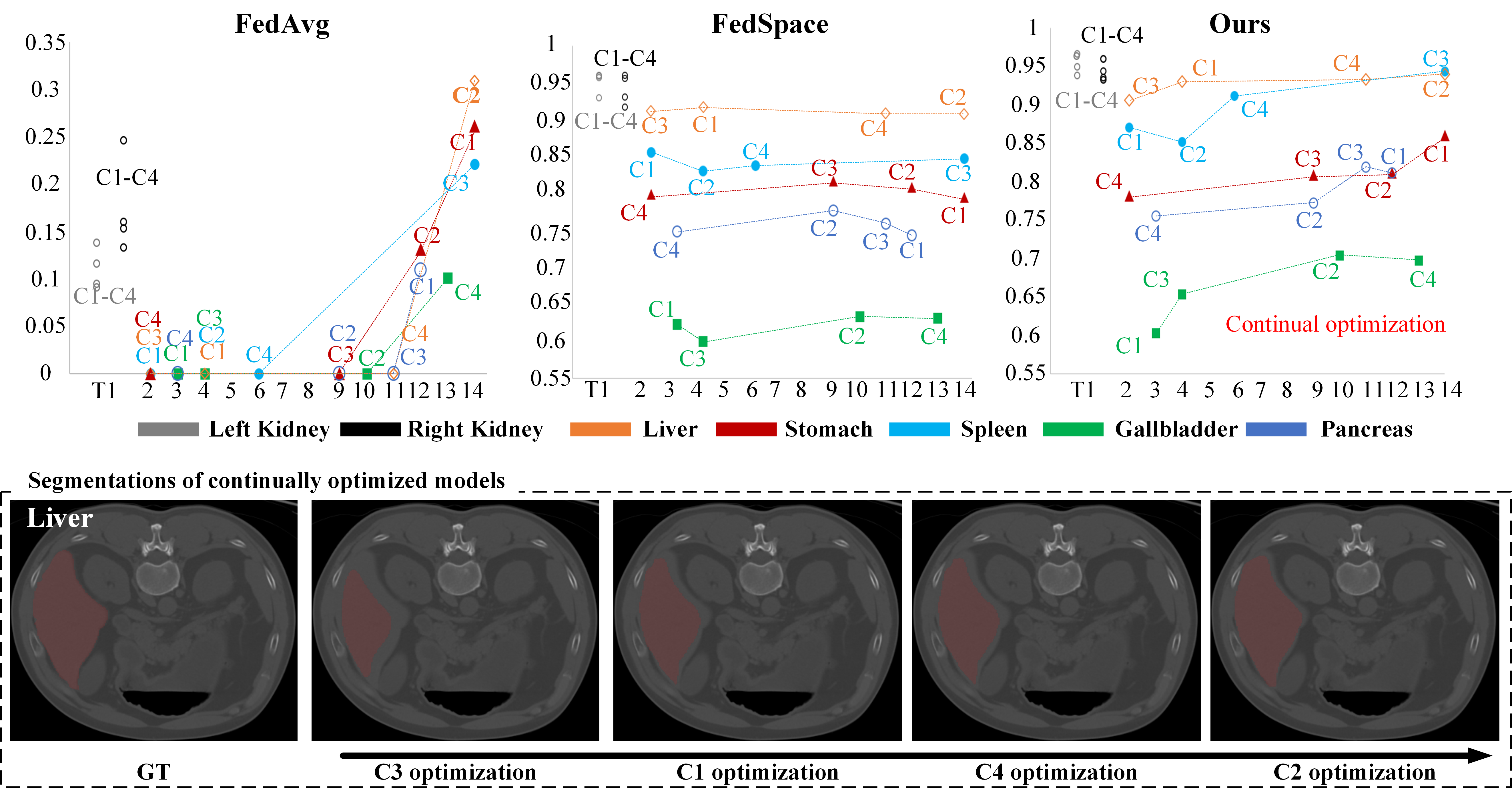}
\caption{FedDAH ensures the continual optimization of different task streams in FCL. The horizontal axis is time step and vertical axis is dice score.}
\label{comdata}
\end{figure*}

The results are also listed in Tab.~\ref{Compare}. It can be found that: (1) In -DAHyper, the proposed DAHyper is replaced by the average strategy in FedAvg\cite{mcmahan2017communication} for each task model and the server preserves all models' parameters. The lower Dice achieved by this configuration indicates that the proposed DAHyper can preserve all model knowledge in server model more effectively. 
(2) In -$\mathcal{L}_{R}$, we remove the regularization of different tasks continual optimizations. It can be found that the performance suffers severe decrease. This indicates that the $\mathcal{L}_{R}$ benefits the FedDAH optimization from historical knowledge. (3) In -$W_{s}$, we remove the recalibration of continual optimization on the same task and just use the uploaded client model to optimize DAHyper. The lower mean Dice demonstrates that the $W_{s}$ can provide effective guidance for the model to optimize the same task model at different time steps.

\subsection{Detailed Analytical Experiments}
\noindent \textbf{Results in Each Continual Step.} To validate the ability of continual learning on different task streams in FCL, we visulaize the learning process of FedDAH, FedSpace, and FedAvg in FCL and the example of continual optimization in our FedDAH learning process (Liver). It can be found that (Fig.~\ref{comdata}): (1) Our FedDAH makes each task achieve the best performance at last optimization. FedDAH gradually improves segmentation at different time steps. This benefit from the task identity is preserved in server to maintain the model better than the previous optimized model with the same task. This indicates that FedDAH ensures the continual optimization on a task at different time steps and different clients. (2) In FedAvg, only the last two task can be optimized and achieves poor performance. In FedSpace, the segmentation performance of pancreas become gradually worse in the learning process. This indicates the knowledge forgetting and optimization bias make it difficult to realize FCL in the real-world. (3) Compare the learning process of the different methods, it could be found that organs suffers unstable optimization in the existing FL and FCL methods. This indicates that the asynchronous tasks makes the existing method hardly work in the real-world FCL. (4) From the visual results of liver, it can be found that the segmentation is gradually improved during the optimization on different clients. This evaluates that our FedDAH could maintain the knowledge in continual learning and correct the optimization bias in FCL.

\begin{table}[!t]
\centering
\caption{The performance of different continual methods on C1 evaluates that FedDAH is able to provide a global FCL model for each sites.}
\scalebox{0.8}{
\begin{tabular}{ccccccccccccccccl}
\hline
Method & lk    & rk    & spl   & sto   & pan   & gal   & liv   & bla   & pro   & aor   & inf   & duo   & eso   & lag   & rag   & AVG   \\ \hline
PLOP   & 0.962 & 0.941 & 0.857 & 0.850 & 0.790 & 0.548 & 0.915 & 0.715 & 0.730 & -     & -     & -     & -     & -     & -     & 0.487 \\
LISMO  & 0.956 & 0.938 & 0.847 & 0.829 & 0.764 & 0.515 & 0.899 & 0.706 & 0.715 & -     & -     & -     & -     & -     & -     & 0.478 \\
CLAMTS & 0.958 & 0.964 & 0.866 & 0.842 & 0.808 & 0.569 & 0.920 & 0.724 & 0.728 & -     & -     & -     & -     & -     & -     & 0.492 \\
CSTSUA & 0.960 & 0.957 & 0.877 & 0.847 & 0.796 & 0.550 & 0.928 & 0.719 & 0.754 & -     & -     & -     & -     & -     & -     & 0.493 \\
FedDAH    & 0.963 & 0.944 & 0.870 & 0.860 & 0.811 & 0.603 & 0.930 & 0.732 & 0.751 & 0.833 & 0.857 & 0.671 & 0.747 & 0.743 & 0.707 & 0.801 \\ \hline
\end{tabular}}
\label{continualmethods}
\end{table}

\vspace{1mm}
\noindent \textbf{Our FedDAH v.s. CL methods.} Training each local model by using CL methods can also be an option to tackle the challenges of different task streams. To evaluate the superiority of FedDAH over CL methods, we compare FedDAH with several advanced CL methods, including PLOP \cite{douillard2021plop}, LISMO \cite{liu2022learning}, CLAMTS \cite{zhang2023continual}, and CSTSUA \cite{10378031}. We train these CL methods on each local data and labels, and the testing is conducted on 15 organs. We take the Client 1 (C1) as an example and the results are shown in Tab.~\ref{continualmethods}. It can be found that: (1) considering that one local client may not see all tasks during train (e.g., C1 does not have the labels of aorta organ), the pure CL methods can not segment these unseen organs (marked as '-' in the table). Instead, our FCL based method FedDAH could makes the partially labeled clients obtain the ability of complete segmentation by learning such knowledge from other clients. (2) Through the comparison of Tab.~\ref{continualmethods} and Tab.~\ref{Compare}, we find that FedDAH can achieve similar performance on Client 1 as it does on other clients. This indicates that our FedDAH could balance the optimization on different clients and share the information among all clients.

\begin{table}[]
\centering
\caption{The details of dataset and the settings of each client using all organs for CL.}
\begin{tabular}{ccccc}
\hline
\multirow{2}{*}{Clients} & \multirow{2}{*}{Num} & \multirow{2}{*}{Task 1(initial)} & \multicolumn{2}{c}{Task 2-14 (random order)}   \\ \cline{4-5} 
                         &                         &                                   & \multicolumn{2}{c}{Shared}                   \\ \hline
C1                       & 125                     &                                   & spleen,          & bladder, prostate,         \\
C2                       & 125                     & left kidney,                      & stomach,pancreas & aorta, inferior vena cava, \\
C3                       & 125                     & right kidney                      & gallbladder,     & duodenum, esophagus,       \\
C4                       & 125                     &                                   & liver            & left, right adrenal gland  \\ \hline
\end{tabular}
\label{datasetr}
\end{table}

\begin{table}[]
\centering
\caption{The performance of each client evaluates the ability of continual learning in our FedDAH.}
\scalebox{0.9}{
\begin{tabular}{ccccccccc}
\hline
\multirow{2}{*}{Task} & \multicolumn{2}{c}{1 Kidney} & \multirow{2}{*}{2} & \multirow{2}{*}{3} & \multirow{2}{*}{4} & \multirow{2}{*}{5} & \multirow{2}{*}{6} & \multirow{2}{*}{7} \\
                      & Left              & Right             &                    &                    &                    &                    &                    &                    \\ \hline
C1                    & 0.963             & 0.944             & Spl:0.870          & Gal:0.603          & Liv:0.930          & Duo:0.671          & Aor:0.833          & Bla:0.732          \\
C2                    & 0.95              & 0.932             & Aor:0.852          & Bla:0.724          & Spl:0.851          & Inf:0.832          & Eso:0.706          & Duo:0.703          \\
C3                    & 0.966             & 0.959             & Liv:0.906          & Aor:0.811          & Gal:0.653          & Pro:0.688          & Eso:0.734          & Inf:0.898          \\
C4                    & 0.938             & 0.936             & Sto:0.781          & Pan:0.755          & Eso:0.693          & Lag:0.703          & Spl:0.911          & Aor:0.883          \\ \hline
                      & 8                 & 9                 & 10                 & 11                 & 12                 & 13                 & 14                 & Avg                   \\ \hline
C1                    & Inf:0.857         & Eso:0.747         & Rag:0.707          & Pro:0.751          & Pan:0.811          & Lag:0.743          & Sto:0.86           &  0.801                  \\
C2                    & Lag:0.694         & Pan:0.772         & Gal:0.704          & Rag:0.703          & Sto:0.811          & Pro:0.816          & Liv:0.939          &   0.799                 \\
C3                    & Rag:0.685         & Sto:0.808         & Duo:0.697          & Pan:0.819          & Bla:0.703          & Lag:0.771          & Spl:0.943          &   0.803                 \\
C4                    & Pro:0.700         & Inf:0.853         & Rag:0.734          & Liv:0.933          & Duo:0.753          & Gal:0.697          & Bla:0.756          &  0.802                  \\ \hline
\end{tabular}}
\label{ModelAnalysiss}
\end{table}

\vspace{1mm}
\noindent \textbf{Performance in Task Level.} To more comprehensively illustrate the superiority of FedDAH, we conduct another FCL setting that each client has seen all the tasks (e.g., organs) during the training, and we show the segmentation performance in the task level. As shown in Tab.~\ref{datasetr}, the four clients still use the left and right kidney segmentation as initialization tasks. Then each client regards the rest 14 organs as task 2 to task 14, but receives the labels in different sequences with random order. We utilize the same test dataset for each time step model for a fair comparison.

The results are listed in Tab.~\ref{ModelAnalysiss}. It indicates the superiority of our FedDAH on real-world FCL with distinct and dynamic task streams. (1) On the shared tasks with the same stream (left and right kidney), all clients can be well optimized with all Dice over 0.9. (2) The same organs at different time steps are continually optimized, such as spleen (`Spl' in the table) is optimized from 0.87 to 0.943.
This evaluates that our FedDAH provides the continual learning ability on the asynchronous task streams. (3) At the same step, different task can be well optimized. Taking step 2 for example, the spleen, aorta, liver, and stomach all have been optimized (0.87, 0.852, 0.906, and 0.781). This evaluates that our FedDAH ensures the different tasks' knowledge preservation in server.

\section{Conclusion}
We propose a novel server-side FCL framework, FedDAH, to enable global knowledge preservation and continually asynchronous task optimization to narrow the gap of deploying FL in real-world application. FedDAH employs a designed hypernetwork to preserve knowledge, incorporates the candidate changes of history, and balances the continual optimization based on similarity. We conduct extensive experiments to validate the effectiveness of our method on the AMOS dataset, outperforming other approaches by a large margin.In the future work, we propose to expand our FedDAH to the scenario of different clients with different organs and tasks in continual learning. This will advance our FedDAH with the ability to eventually train foundation model that is compatible with existing medical foundation models releasing from the data collection.

\section{Acknowledge}
This work was supported by Ministry of Education Tier 1 Start up grant, NUS, Singapore (A-8001267-01-00); Ministry of Education Tier 1 grant, NUS, Singapore (A-8001946-00-00); and the National Natural Science Foundation of China (Grant No. 82441021).


%
%
%
\bibliographystyle{splncs04}
\bibliography{samplepaper}

\end{document}